# Applications of Artificial Intelligence for Cross-language Intelligibility Assessment of Dysarthric Speech

Eunjung Yeo[1], Julie M. Liss[2], Visar Berisha[2], David R. Mortensen[1]

Carnegie Mellon University[1], Arizona State University[2]

**Abstract**

**Purpose**: Speech intelligibility is a critical outcome in the assessment and management of dysarthria, yet most research and clinical practices have focused on English, limiting their applicability across languages. This commentary introduces a conceptual framework—and a demonstration of how it can be implemented—leveraging artificial intelligence (AI) to advance cross-language intelligibility assessment of dysarthric speech.

**Method**: We propose a two-tiered conceptual framework consisting of a universal speech model that encodes dysarthric speech into acoustic-phonetic representations, followed by a language-specific intelligibility assessment model that interprets these representations within the phonological or prosodic structures of the target language. We further identify barriers to cross-language intelligibility assessment of dysarthric speech, including data scarcity, annotation complexity, and limited linguistic insights into dysarthric speech, and outline potential AI-driven solutions to overcome these challenges.

**Conclusion**: Advancing cross-language intelligibility assessment of dysarthric speech necessitates models that are both efficient and scalable, yet constrained by linguistic rules to ensure accurate and language-sensitive assessment. Recent advances in AI provide the foundational tools to support this integration, shaping future directions toward generalizable and linguistically informed assessment frameworks.

**Introduction**

Dysarthria is a motor speech disorder resulting from neuromuscular impairments, affecting key components of speech production such as respiration, phonation, resonance, articulation, and prosody (Darley et al., 1969, 1975; Enderby, 1980). These disruptions manifest as acoustic-perceptual speech characteristics that deviate from those of healthy speakers, often resulting in reduced intelligibility, communication breakdowns, and a diminished quality of life (Piacentini et al., 2014; Spencer et al., 2020). Accurate assessment of speech intelligibility is therefore crucial in clinical practices, including diagnosis, monitoring, and intervention strategy development. Given the rising global burden of neurological conditions associated with dysarthria (World Health Organization, 2024), there is a need for assessment methods that are effective across languages. We posit that advances in artificial intelligence (AI) offer promising avenues for cross-language intelligibility assessment of dysarthric speech.

The role of native language in motor speech disorders has received relatively little attention in speech-language pathology, likely due to the assumption that speech motor control is language-universal. Yet, emerging research challenges this view as it pertains to speech intelligibility (Miller & Lowit, 2014; Pinto et al., 2017; Kim et al., 2017). Variations in phonological contrasts and rhythmic patterns across languages can significantly alter how speech impairments contribute to intelligibility degradation. This growing recognition is reflected in a special issue of the Journal of Speech, Language, and Hearing Research titled "Native Language, Dialect, and Foreign Accent in Dysarthria: Clinical and Research Considerations" (Kim, 2024). The collection underscores the importance of considering the native languages of both speakers and listeners in dysarthric speech intelligibility assessment and intervention for optimal clinical outcomes.

Perceptual evaluation, the conventional method for assessing speech intelligibility in dysarthria, presents inherent limitations for cross-language assessment due to its subjectivity (Hirsch et al., 2022) and perceptual bias (Kim et al., 2024). In particular, a mismatch between the native languages of the speech language pathologist (SLP) and the patient can bias assessments—overlooking features critical in the patient's language while overemphasizing those salient in their own. For example, an English-speaking SLP may underestimate the impact of monotonicity in tonal languages such as Mandarin, where pitch contrasts are essential for distinguishing word meaning. Conversely, the same clinician may overemphasize vowel precision when assessing Spanish, where the smaller vowel inventory results in lower vowel confusability compared to English. These linguistic mismatches introduce bias into clinical assessments and undermine intelligibility assessment accuracy. Such limitations highlight the need for an objective, language-sensitive framework for cross-language intelligibility assessment of dysarthric speech.

Recent advances in AI offer timely opportunities to address these limitations in cross-language intelligibility assessment of dysarthric speech. This paper proposes a conceptual framework consisting of two components: (1) a *universal speech model* that encodes the acoustic-phonetic manifestations of dysarthric speech independently of linguistic context, followed by (2) a *language-specific intelligibility assessment model* that interprets these representations in relation to linguistic rules of the target language. By aligning generalizable acoustic-phonetic representations with linguistically-informed adaptation, this framework aims to deliver objective, and bias-mitigated intelligibility assessments applicable across languages.

The remainder of this commentary is organized as follows. It begins with a review of recent studies on cross-language intelligibility assessment of dysarthric speech, presenting findings that offer insights into the need for language-sensitive approaches. The proposed framework is then introduced, integrating a universal speech model with a language-specific intelligibility assessment model. Subsequently, linguistic factors relevant to dysarthric speech and intelligibility are explored for their potential to inform language-specific intelligibility modeling. This is followed by a discussion of the barriers that have hindered progress in cross-language intelligibility assessment and AI-driven strategies to address these challenges. The specific instantiation of the framework is then presented to demonstrate the practical implementation of the proposed framework. Finally, the commentary concludes with reflections on future directions for advancing cross-language intelligibility assessment of dysarthric speech.

**Recent Studies on Cross-Language Intelligibility Assessment of Dysarthric Speech**

Dysarthria manifests through acoustic-phonetic deviations from general populations, including imprecise articulation, reduced vowel space, hypernasality, monopitch, rhythmic disturbances, and atypical pausing patterns (Darley et al., 1969, 1975; Enderby, 1980). Although these disruptions form a broadly consistent profile across languages, their perceptual impact is shaped by the phonological and prosodic structures of the target language. This has motivated cross-language studies examining how acoustic-phonetic deviations interact with linguistic systems.

Liss et al. (2013) demonstrated that, in American English (AE), different subtypes of dysarthria–each with different segmental and suprasegmental degradation patterns–differentially impact a listener's ability to segment the dysarthric speech stream into words. Because of the statistical probabilities of the English language, listeners will be most successful in lexical segmentation of degraded speech when they treat strong syllables (those produced with relative emphasis) as word-onsets (Cutler & Norris, 1988). Thus, the preservation of syllable contrastivity (relative differences in duration, loudness, vowel quality) is necessary for employing this perceptual strategy. The different dysarthria subtypes affected syllable contrastivity in distinct ways, and listeners' patterns of lexical boundary errors (LBEs) reflected how the contrastive cues were degraded. This study exemplifies the interaction between language-specific perceptual strategies and dysarthria manifestations, offering a template for conceptualizing cross-language intelligibility assessment.

This interaction was further supported by Kim and Choi (2017), who analyzed Parkinson's disease (PD) speech in AE and Korean. Their analysis of acoustic vowel space, voice onset time (VOT) contrast scores, normalized pairwise variability index, and articulation rate identified VOT as a language-specific acoustic variable predictive of intelligibility in Korean. The authors attributed this to the denser VOT space in Korean, which features a three-way stop contrast, in contrast to the two-way contrast found in English. Their findings confirmed that universal motor deficits interact with phonological systems, shaping how intelligibility deteriorates across languages.

Pinto et al. (2017) expanded this perspective, proposing that intelligibility deficits in hypokinetic dysarthria arise from the interaction between motor impairments and language-specific phonological and prosodic structures. They argued that both segmental features, such as VOT contrasts, and suprasegmental features, such as pitch prominence, modulate how motor symptoms affect intelligibility across languages. For example, reduced pitch variation may impair French more severely, where prominence is marked by pitch, than Portuguese, where prominence relies more on vowel reduction and duration. This framework emphasizes that intelligibility assessment must account for these linguistic structures, as motor symptoms alone do not fully explain cross-language variability in dysarthria.

Despite this recognition, empirical studies have prioritized identifying language-robust acoustic features to support scalable assessment tools. Favaro et al. (2023a) and Kováč et al. (2024) applied machine learning across five different languages, consistently identifying F0 variability and pause patterns as reliable markers of dysarthria. Pinto et al. (2024), focusing on French and Portuguese, identified respiratory and rate-related features, including maximum phonation time and diadochokinesis (DDK) rate, as stable predictors of PD dysarthria across these languages. However, the cross-linguistic generalizability of pitch-related features remains contested. While Favaro et al. (2023a) and Kováč et al. (2024) reported monopitch as a consistent marker across languages, Pinto et al. (2024) found that pitch cues were not among the most informative acoustic variables for distinguishing PD from controls, particularly in Portuguese. This discrepancy underscores that so-called language-robust features may in fact be conditioned by the languages under study.

Moreover, reliance on language-robust features alone has proven insufficient for cross-linguistic intelligibility prediction. Kováč et al. (2024) found that models trained on combined data from multiple languages using identified language-robust features performed worse than models trained separately for each individual language (monolingual setting). Yeo et al. (2022)

further demonstrated that incorporating language-specific features improved model generalization across AE, Korean, and Tamil, in comparison to monolingual settings. These findings reinforce Pinto et al.'s (2017) framework, affirming that intelligibility loss in dysarthria reflects the interaction between motor impairments and linguistic systems.

In summary, although dysarthric speakers exhibit similar physical speech motor impairments across languages, the communicative challenges they face differ by native language. Therefore, cross-language intelligibility assessment of dysarthric speech requires considerations on two dimensions: shared acoustic-phonetic manifestations of dysarthria and language-specific phonological and prosodic structures that shape intelligibility. Features related to respiration and speech rate may function as language-robust markers, whereas those linked to phonemic contrasts, prosody, and rhythm require language-sensitive interpretation. These findings collectively point to the need for assessment frameworks that can accommodate both generalizable acoustic modeling and language-specific linguistic constraints.

**Conceptual framework for cross-language intelligibility assessment with AI**

We propose a two-tiered AI-based framework for cross-language intelligibility assessment of dysarthric speech (Figure 1). The framework decouples acoustic-phonetic encoding from linguistic interpretation by structuring them into two sequential components: a universal speech model, which encodes observable speech patterns without reference to linguistic norms, and a language-specific intelligibility assessment model, which interprets these patterns constrained by linguistic knowledge.

The *universal speech model* functions as the initial encoder, transforming dysarthric speech into language-independent representations that capture its core acoustic-phonetic characteristics. These representations may take the form of phone sequences or acoustic features such as pitch contours, segmental durations, and spectral properties. This component produces generalizable descriptors of dysarthric speech that can be extracted from speech across different languages. This architecture reduces reliance on language-specific training data and supports a unified encoding approach applicable to diverse linguistic contexts.

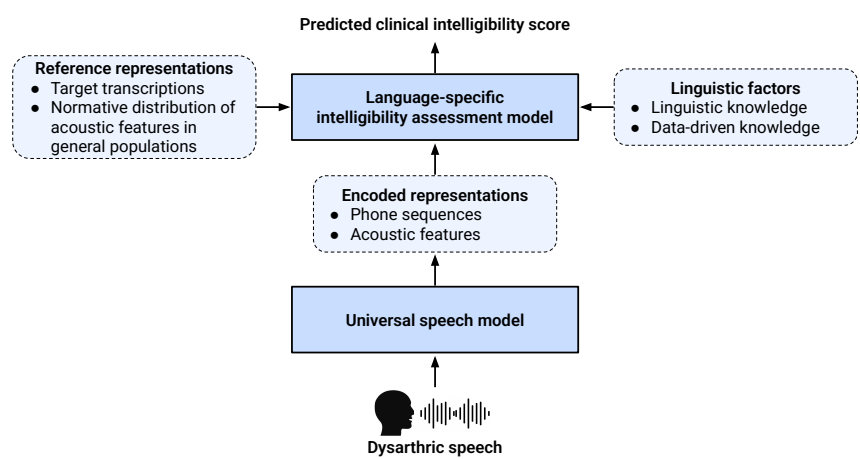

Figure 1. Conceptual framework for cross-language intelligibility assessment.

The *language-specific intelligibility assessment model* performs the interpretive task. This component draws from three sources: the **encoded representations** generated by the universal model; **reference representations** such as target transcriptions or normative distribution of acoustic features from general populations; and **linguistic factors**. These factors include both expert-defined linguistic knowledge, such as phoneme inventories, phonological rules, and rhythmic typologies—and data-driven insights derived from healthy and dysarthric speech corpora. Further details on linguistic factors will be demonstrated in subsequent sections. Intelligibility is quantified by comparing the encoded and reference representations using distance-based metrics such as word error rate (WER), phoneme error rate (PER), and acoustic deviation scores like Goodness of Pronunciation (Witt & Young, 2000). Linguistic factors constrain these comparisons by weighting or filtering contrasts based on the linguistic rules of the target language.

By decoupling acoustic encoding from linguistic interpretation, the proposed framework enables intelligibility assessment that balances cross-linguistic scalability with language-specific precision. The universal speech model generates shared representations of dysarthric speech applicable across languages, while the language-specific assessment model imposes lightweight, interpretable constraints informed by the structure of the target language. This modular design reduces the need to retrain full models for each language, lowering resource demands and enhancing clinical applicability across languages.

**Potential linguistic factors in cross-language intelligibility assessment of dysarthric speech**

Cross-language intelligibility assessment of dysarthric speech must consider both shared acoustic-phonetic manifestations and language-specific linguistic structures that shape how these features affect intelligibility. This section outlines four potential linguistic factors that differentially influence intelligibility across languages (see Table 1). These factors correspond to the *linguistic factors* component in the proposed framework; all four can be inferred from established linguistic principles and further complemented by data-driven approaches.

Tonal contrast is critical in tonal languages where pitch contours signal lexical distinctions (Huang & Johnson, 2011). For example, in Mandarin Chinese, the syllable "ma" varies meaning by tone: [ma˧˩˧](马) means " horse," while [ma˥˩] (骂) means "to scold." Consequently, dysarthria-related monotonicity decreases speech intelligibility in tonal languages but not in non-tonal languages like English.

Rhythmic typology stress-timed (English), syllable-timed (Spanish), and mora-timed (Japanese) also modulates intelligibility (Nespor et al., 2011). Stress-timed languages juxtapose stressed and unstressed syllables, whereas syllable-timed languages exhibit relatively uniform syllable durations. Mora-timed languages structure rhythm around mora–a timing unit assigned to short vowels or light syllables–and each mora tends to have roughly equal duration (Hayes, 1989). Given these typological differences, dysarthria-related symptoms of producing equal and even stress/duration of syllables in ataxic dysarthria may have a greater impact on English than Spanish intelligibility. In contrast, hyperkinetic dysarthria, characterized by irregular rhythmic patterns, can reduce intelligibility across languages.

Phonetic variation highlights the density of a language's acoustic space (Kim & Choi, 2017; Yeo et al., 2023b). Languages like English with dense vowel space require greater articulatory precision to maintain phonemic contrasts and prevent confusion with neighboring phonemes. Thus dysarthria-related vowel centralization may be more detrimental for English than Spanish intelligibility, which has a simpler vowel system. Conversely, Spanish fricatives occupy a less crowded acoustic space due to fewer voicing and place contrasts, making them less prone to confusability than English fricatives. These examples illustrate that the same articulatory deviation can have different intelligibility consequences depending on the language's phonetic landscape.

Phonological contrast shapes how pronunciation deviations are classified across languages. For instance, substituting /ɾ/ with /r/ in Spanish (e.g., [peɾo] "but" → [pero] "dog") constitutes a substitution error due to their phonemic contrast. In English, however, the same substitution would be interpreted as a distortion, since /ɾ/ is not contrastive. Allophonic variation also plays a role. In English, substituting /s/ with /z/ (e.g., sip → zip) represents a phonemic error, while in Colombian Spanish, [z] may surface as an allophone of /s/ before voiced stops (e.g., desde → [dez.de]), and such voicing assimilation is optional and not typically considered erroneous. Thus, the same phonetic deviation may be interpreted as a significant intelligibility error in one language but as permissible variation in another.

**Table 1. Potential linguistic factors in cross-language intelligibility assessment.**

| Linguistic factors | Related Speech Impairment | Cross-Language Difference | Impact of dysarthria |
|---|---|---|---|
| Tonal contrast | Irregularity in pitch | Tonal language vs non-tonal language | Dysarthric speakers in tonal languages are expected to require greater precision in producing pitch contours for lexical differentiation, whereas pitch irregularity is less critical in non-tonal languages. |
| Rhythmic typology | Rhythmic disruption | Stress-timed vs syllable-timed vs mora-timed | Hypokinetic dysarthria may hinder speech intelligibility in stress-timed languages, while hyperkinetic dysarthria can disrupt rhythmic patterns across all typologies. |
| Phonetic variation | Imprecise articulation | Vowel Space Area (VSA) | Due to a denser VSA in English compared to Korean and Spanish, English speakers are more likely to experience greater vowel confusability than Korean and Spanish speakers. |
| | Imprecise articulation | Acoustic density | Spanish has a smaller fricative inventory than English, with limited voicing and place contrasts, resulting in lower acoustic density and potentially less confusability in fricatives. |
| Phonological contrast | Imprecise articulation | Phoneme inventory | Errors like /n/ to /ɲ/ are substitution errors in Spanish but are perceived as distortion errors in English due to the absence of /ɲ/ in its phoneme inventory. |
| | Imprecise articulation | Allophonic variations | /s/ to /z/ substitution constitutes a phonemic error in English, but may be considered acceptable in Colombian Spanish, where [z] functions as an allophonic variant of /s/ before voiced stops. |

**Barriers and Solutions to Cross-language Intelligibility Assessment of Dysarthric Speech**

Cross-language intelligibility assessment of dysarthric speech faces challenges that hinder progress in both perceptual and AI-based evaluation. For AI-based evaluation, one major barrier is data scarcity, as large, diverse datasets are essential for building AI models that generalize across languages. In perceptual evaluation, annotation complexity presents difficulties, as intelligibility judgments often require linguistically-informed annotations and cross-linguistic sensitivity. A third barrier, limited linguistic insight into dysarthric speech, affects both domains, constraining the development of language-sensitive frameworks. In this section, we outline these challenges and propose corresponding solutions that integrate advances in AI.

**Table 2. Potential Applications of AI to overcome barriers for cross-language intelligibility assessment**

| Limiting factors | Method | Description |
| --- | --- | --- |
| Data scarcity | Voice Conversion | Augment training data by generating healthy speech and dysarthric-like speech. |
| | Text-to-Speech | |
| | Transfer learning | Transfer classifiers trained on high-resource languages to low-resource languages. |
| | Cross-lingual Self-Supervised Learning (SSL) Models | Leverage pre-trained cross-lingual SSL models, which perform effectively on downstream tasks with smaller labeled data compared to traditional supervised methods. |
| Complexity of Annotations | Automatic Speech Recognition (ASR) | Use ASR to generate word-level transcriptions, reducing the need for manual transcriptions. |
| | Universal Phone Recognition Model (UPRM) | Use UPRM to generate phonetic transcriptions, reducing the need for manual transcriptions. |
| | Acoustic feature extraction systems | Utilize tools to extract acoustic features, enabling efficient and objective extraction of relevant characteristics. |
| Limitations in Linguistic Insights | Data-driven analysis | Leverage data-driven approaches to uncover linguistic characteristics and the language-specific relationship between speech impairment and intelligibility. |

*Data scarcity*

The first barrier is the limited availability of open-source multilingual speech corpora, particularly those including dysarthric speech. In particular, while several well-established corpora exist for English—covering both healthy and dysarthric speech—comparable resources for other languages remain scarce. Even among healthy speech datasets, many lack the linguistic diversity required for universal speech modeling, constraining the capacity to generalize across languages. For dysarthric speech, this scarcity is even more pronounced outside English, with few accessible corpora to support cross-linguistic model development (Bhat & Strik, 2025). Until recently, deep learning-based models were highly data-dependent, performing poorly in the absence of large, annotated corpora (Baevski et al., 2020). This limited the development of intelligibility assessment models in different languages.

Given that collecting speech data is both labor- and time-intensive, synthetic data generation offers a practical alternative. For healthy speech, text-to-speech (TTS) and voice conversion (VC) technologies can produce multilingual synthetic datasets, expanding language coverage where resources are insufficient (Kumar et al., 2020). In the dysarthric speech domain, generative models can simulate impairments. VC systems have been applied to transform healthy into dysarthric-like speech (Jiao et al., 2018; Jin et al., 2023; Wang et al., 2023), while TTS systems have been adapted to synthesize speech reflecting the acoustic characteristics of dysarthria from text (Hu et al., 2023; Leung et al., 2024). These methods enhance both the diversity of healthy speech used in universal modeling and the capacity for language-specific adaptation by enabling the generation of dysarthric speech across languages.

Complementing synthetic data generation, recent AI methodologies have become increasingly data-efficient, reducing reliance on large annotated datasets. Transfer learning allows models pre-trained on high-resource languages—those with extensive annotated speech corpora, such as English—to be adapted to low-resource languages where such resources are limited or unavailable (Bhat & Strik, 2020; Vásquez-Correa, 2021). This approach is generally most effective when the source (high-resource) and target (low-resource) languages are closely related, as in the case of Dutch and Flemish. Here, resource status is typically defined by the availability of annotated speech data. However, the specific linguistic characteristics that support successful cross-language transfer remain an open research question.

Self-supervised learning (SSL) models (Baevski et al., 2020) mitigate data scarcity by leveraging pre-training on large-scale unlabeled multilingual corpora to learn generalizable acoustic-phonetic representations (Pasad et al., 2021; Choi et al., 2024). These models offer two advantages: (1) SSL models achieve strong performance with small labeled data through fine-tuning (Baevski et al., 2020), and (2) SSL representations can be directly used as input features for classification or regression models. Recent studies have demonstrated SSL representations outperforming hand-crafted features on intelligibility assessment (Yeo et al., 2023a; Favaro et al., 2023b; Javanmardi et al., 2024). For cross-language intelligibility assessment, cross-lingual SSL models such as XLSR-53 (Conneau et al., 2021) and XLS-R (Babu et al., 2022) are well-suited, as they are pre-trained on multilingual speech and provide more robust and transferable acoustic-phonetic representations than monolingual models.

*Complexity of Annotations*

Manual transcription and rating remain central to perceptual evaluation of dysarthric speech intelligibility but pose significant challenges, particularly in cross-linguistic contexts. Transcribing dysarthric speech requires attention to both speech impairments and language-specific structures, making the process labor-intensive and prone to inconsistencies, which weakens the foundation for reliable intelligibility assessment. Outputs from AI models can serve as first-pass transcriptions, where they provide a practical starting point for human annotators and reduce the overall annotation burden.

Automatic speech recognitions (ASRs) have been explored as scalable alternatives to manual transcription. Word error rate (WER) derived from ASR output has shown strong correlations with clinician-rated intelligibility scores (Karbasi & Kolossa, 2022; Gutz et al., 2023), providing an objective proxy for manual transcription. Recent multilingual ASR systems, such as Whisper (Radford et al., 2023), WhisperX (Bain, 2023), XLSR-53 (Conneau et al., 2021), XLS-R (Babu et al., 2022), and Massively Multilingual Speech (Pratap et al., 2024), extend this potential across languages, reducing annotation burden and mitigating perceptual variability. Nevertheless, word-level transcriptions are limited in capturing fine-grained phoneme-level deviations often critical to intelligibility, as ASR systems prioritize plausible lexical sequences over phonetic accuracy.

Universal phoneme recognition models (UPRMs), such as Allosaurus (Li et al., 2020), Wav2Vec2Phoneme (Xu et al., 2022), and AlloPhant (Glocker et al., 2023), address this limitation by generating phone-level transcriptions independent of language models. When aligned with canonical phoneme sequences generated from multilingual grapheme-to-phoneme (G2P) systems

(Mortensen et al., 2018), these outputs enable computation of interpretable phoneme-level metrics such as PER and categorized error types (e.g., insertion, substitutions, deletions, distortions).

In parallel, acoustic feature extraction systems—such as eGeMAPS (Eyben et al., 2016) and DisVoice (Vásquez-Correa et al., 2018)—offer efficient and objective extraction of relevant characteristics. For instance, eGeMAPS has been explored for intelligibility assessment in Dutch dysarthric speech (Xue et al., 2019), and DisVoice has supported severity classification in English (Joshy et al., 2023). However, recent studies report substantial intra-speaker variability and limited clinical robustness (Stegmann et al., 2020), highlighting the need for further refinement to ensure broader clinical applicability.

*Limited linguistic insights into dysarthric speech*

Linguistic analysis of dysarthric speech remains limited across languages, posing challenges for developing cross-language intelligibility assessment tools for both perceptual and AI-based evaluations. Most prior research has focused on English, leaving underexplored how dysarthric speech interacts with phonological and prosodic systems of other languages (Pinto et al., 2017; García et al., 2023). Extracting such linguistic insights traditionally requires expert-driven phonetic and phonological analysis, which is resource-intensive and difficult to scale.

Data-driven analyses offer a scalable alternative. Recent studies show that SSL representations encode multi-level linguistic information, particularly phonetic and phonological information (Choi et al., 2024; 2025). For example, Choi et al. (2025) found that phoneme-level SSL representations cluster by phoneme, with allophones emerging as subclusters. This structure enables the derivation of language-specific acoustic spaces for each phoneme from general populations. These distributions align with linguistic factors such as phonetic variation introduced earlier. Ramesh et al. (2021) demonstrated that SSL models encode phonotactic representations relevant for language identification, suggesting their utility to capture language-specific sequencing constraints. Collectively, these findings underscore the potential of data-driven methods to reveal linguistic patterns that constrain the interpretation of acoustic-phonetic deviations in intelligibility assessment.

Interpretability techniques also represent a form of data-driven analysis. Methods such as Shapley value-based feature attribution (Kováč et al., 2024) and attention-based mechanisms (Choi et al., 2025) support the development of language-specific intelligibility modules by identifying which acoustic-phonetic deviations are most predictive of intelligibility in a given language. These techniques allow models to filter out irrelevant features and assign greater weight to linguistically salient deviations.

**Specific Instantiation of the Proposed Framework**

Figure 2 presents a specific instantiation of the proposed framework, applied to a dysarthric speech sample in Spanish, exemplified by the utterance piel salud libro bar ("skin health book bar"). First, the speech is processed by Wav2Vec2Phoneme (Xu et al., 2022), an UPRM instance that outputs phone sequences independent of the spoken language. These recognized phone sequences are then aligned with a target transcription derived from a G2P model.

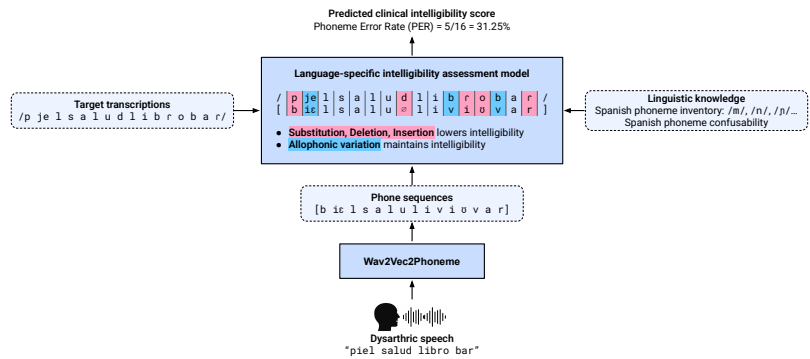

**Figure 2. Specific instantiation of the proposed framework.**

The alignment can be refined by integrating linguistic factors, the language-specific phoneme confusability, derived from acoustic spaces learned through data-driven analysis. For instance, Spanish lacks the /ɔ/ vowel and uses only /o/, whereas English includes both /oʊ/ and /ɔ/, resulting in greater acoustic proximity and higher confusability between these vowels in English. By incorporating empirically derived phoneme confusion matrices, the framework enables language-specific adjustments to alignment, improving its sensitivity to acoustic-phonetic realities across languages.

After alignment, language-specific phonological knowledge, in this case the Spanish phoneme inventory, is incorporated to refine error evaluation during intelligibility scoring. For example, in Spanish, [b] and [v] are allophonic variants and should not be penalized as substitution errors. Without applying phonological constraints, the PER for this utterance is 8 errors out of 16 phones (50%). After accounting for these language-specific corrections, the PER decreases to 5 errors (31.25%).

**Conclusion**

Cross-language intelligibility assessment of dysarthric speech remains a critical yet underdeveloped area. Effective assessment requires models that balance scalability with language-sensitive interpretation by combining universal acoustic-phonetic modeling and linguistically informed adaptation. This commentary introduced a two-tiered conceptual framework that addresses these dual components and illustrated how recent advances in AI can facilitate scalable and language-informed assessment. Continued research is essential to refine and validate these approaches, ensuring their consistency with perceptual evaluation standards and responsiveness to cross-linguistic variation.

**Data Availability Statement** No datasets were created for this commentary.

**Artificial Intelligence Statement** ChatGPT 4-o (OpenAI, 2024) was used to improve the clarity and readability of the manuscript draft. The resulting text was revised by the authors prior to inclusion in the manuscript.